\documentclass{article}

\usepackage{arxiv}
\usepackage{amsmath}
\usepackage[utf8]{inputenc} 
\usepackage[T1]{fontenc}    
\usepackage{hyperref}       
\usepackage{url}            
\usepackage{booktabs}       
\usepackage{amsfonts}       
\usepackage{nicefrac}       
\usepackage{microtype}      
\usepackage{lipsum}
\usepackage{graphicx}
\graphicspath{ {./images/} }
\usepackage{booktabs}
\usepackage{xcolor}
\usepackage{amssymb}
\usepackage{subcaption}
\usepackage{verbatim}
\usepackage[normalem]{ulem}
\usepackage{comment}

\date{}
\title{A General Algorithm for Detecting Higher-Order Interactions via Random Sequential Additions}

\author{
 Ahmad Shamail \\
  Department of Computer Science\\
  University of Arizona\\
  Tucson, AZ 85721 \\
  \texttt{shamail@arizona.edu} \\
   \And
 Claire McWhite \\
  Department of Molecular and Cellular Biology\\
  University of Arizona\\
  Tucson, AZ 85721 \\
  \texttt{clairemcwhite@arizona.edu} \\
}

\begin{document}
\newcommand{\ahmad}[1]{\textcolor{red}{[<Ahmad> #1]}}
\newcommand{\ans}[1]{\textcolor{cyan}{[<Answer> #1]}}
\newcommand{\claire}[1]{\textcolor{blue}{[<Claire> #1]}}

\newcommand{\indep}{\perp\!\!\!\perp}
\newcommand{\nindep}{\not\!\perp}

\maketitle
\begin{abstract}


Many systems exhibit complex interactions between their components: some features or actions amplify each other's effects, others provide redundant information, and some contribute independently. We present a simple geometric method for discovering interactions and redundancies: when elements are added in random sequential orders and their contributions plotted over many trials, characteristic L-shaped patterns emerge that directly reflect interaction structure. The approach quantifies how the contribution of each element depends on those added before it, revealing patterns that distinguish interaction, independence, and redundancy on a unified scale. When pairwise contributions are visualized as two--dimensional point clouds, redundant pairs form L--shaped patterns where only the first-added element contributes, while synergistic pairs form L--shaped patterns where only elements contribute together. Independent elements show order--invariant distributions. We formalize this with the L--score, a continuous measure ranging from $-1$ (perfect synergy, e.g. $Y=X_1X_2$) to $0$ (independence) to $+1$ (perfect redundancy, $X_1 \approx X_2$). The relative scaling of the L--shaped arms reveals feature dominance in which element consistently provides more information. Although computed only from pairwise measurements, higher--order interactions among three or more elements emerge naturally through consistent cross--pair relationships (e.g. AB, AC, BC). The method is metric--agnostic and broadly applicable to any domain where performance can be evaluated incrementally over non-repeating element sequences, providing a unified geometric approach to uncovering interaction structure.

\end{abstract}

\section{Introduction}

Understanding how elements jointly influence system behavior remains a core challenge in modeling and learning. Even when individual effects are measurable, interactions between elements that influence an outcome are often difficult to disentangle. Detecting both redundancy and synergy among features is essential for both interpreting and improving complex models.

In machine learning, this challenge arises when multiple features contribute to a prediction in overlapping or synergistic ways. Two features may appear important in isolation yet redundant together, or individually weak but strongly predictive in combination. Similar dynamics occur in reinforcement learning and control settings,  where discrete actions may overlap in effect or yield improvement only when combined in specific sequences.

We present a simple way to uncover these relationships by observing how system performance changes as elements are added in random sequential orders. Across many such trials, each element’s incremental contribution depends on those added before it. When pairwise contribution values are plotted as scatter points, redundant and interacting pairs form distinctive geometric clouds. Redundant pairs produce L-shaped point distributions in which the first-added element contributes strongly while the second adds little (Figure 1A). Synergistic pairs form mirror-oriented L-shapes, where performance improves primarily when both elements are included (Figure 1B). These patterns arise directly from performance dynamics, without assumptions about the underlying model or domain. The relative scaling of the two arms of the L further reflects dominance or asymmetry between elements.

Our framework quantifies these relationships using the L-score, a continuous measure ranging from $-1$ (perfect synergy, e.g. $Y=X_1X_2$) to $0$ (independence) to $+1$ (perfect redundancy, $X_1\approx X_2$). Although computed only from pairwise measurements, consistent cross--pair patterns expose higher--order interactions among three or more elements. The approach is metric--agnostic and broadly applicable to any domain where performance can be evaluated incrementally and outcomes depend on discrete, potentially interdependent components.

\begin{figure}
    \centering
    \includegraphics[width=1\linewidth]{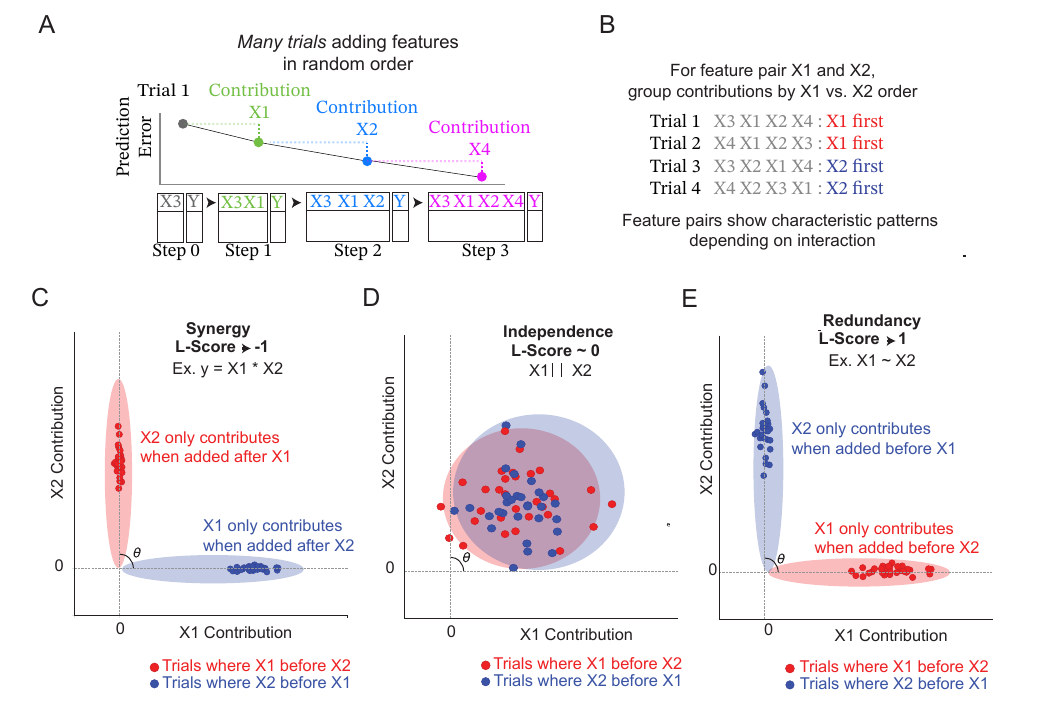}
    \caption{Order-dependent feature contributions reveal redundancy and synergy.
    (A) Features are added in random sequential orders across trials, and marginal contributions are measured via performance change.
    (B) For each feature pair $(X_1, X_2)$, contributions are grouped by relative addition order.
    (C) Synergy: mirror L-shaped distributions where features contribute primarily when added after one another ($L \approx -1$).
    (D) Independence: diffuse, symmetric contribution clouds with no order dependence ($L \approx 0$).
    (E) Redundancy: L-shaped distributions where only the first-added feature contributes ($L \approx +1$).}
    \label{fig:Figure1}
\end{figure}

Many techniques have been developed to detect synergy and redundancy between features in complex models. For detecting synergistic interactions, variance-based methods like Friedman's $H$-statistic \cite{friedman2008predictive}, measure the portion of model output variance that arises from feature interactions rather than additive effects. Sobol indices \cite{owen2013higherordersobolindices} perform global sensitivity analysis by decomposing total variance into contributions from individual features (first-order) and their interactions (higher-order indices). When features act synergistically, their interaction indices are large, indicating their combined effect exceeds the sum of their individual effects. Visualization methods, including Partial Dependence and Accumulated Local Effects plots \cite{apley2019visualizingeffectspredictorvariables}, provide qualitative insight into nonlinear dependencies between features. The widely used SHAP interaction values and its extensions \cite{lundberg2017unifiedapproachinterpretingmodel,dhamdhere2020shapleytaylorinteractionindex,tsai2023faithshapfaithfulshapleyinteraction} provide a principled framework for quantifying pairwise synergistic interactions by measuring how one feature's contribution changes in the presence of another, with the magnitude indicating interaction strength and the sign indicating whether features amplify or diminish each other's effects. For detecting redundancy, information--theoretic approaches use mutual information to quantify shared information between features, while methods like variance inflation factors and feature clustering identify groups of interchangeable predictors.  Together, these methods provide complementary perspectives on feature relationships. Despite these various approaches, distinguishing redundancy from synergy typically requires different measures, making unified analysis challenging, with no single metric identifying both redundancy and synergy on a single scale. Additionally, most methods have increasing complexity for measuring higher-order interactions beyond pairwise. 

Our method contributes to this landscape by introducing a framework that (1) quantifies synergy and redundancy on a continuous scale  in $[-1,1]$, (ii) yields a dominance coefficient derived from the geometric scaling of contribution arms, and (iii) uncovers higher-order dependencies through consistent cross--pair relationships without requiring explicit $k$--way terms.

\section{Our Method: Sequential Feature Addition and the L-Score}

We introduce the \textbf{L-score}, a novel metric that captures both redundancy and dependency within a single, interpretable value. The L-score is computed based on performance gains during randomized sequential feature additions. Across many such permutations, certain performance patterns emerge for pairs or groups of related features. These patterns are captured geometrically by analyzing point clouds formed by plotting the performance contribution of feature pairs 

\subsection*{Sequential Feature Addition and the L-Pattern Visualization}

For a set of features, we perform multiple trials where one feature is added to the model first, followed by each additional feature in a random order. At each step, we measure the marginal reduction in mean squared error (MSE) contributed by the added feature (Figure \ref{fig:Figure1}A). For a pair of features, we then class each trial by which feature occurred first  (Figure \ref{fig:Figure1}B).

This creates a pair of color-coded clouds, whose shape and orientation convey structural information about the relationship between the features :

When a pair of features interact synergistically in predicting the target (i.e., $Y = X_1 \cdot X_2$), each feature is  only capable of substantially contributing to the reduction of MSE when both features are present (Figure \ref{fig:Figure1}C). This creates a characterize L-pattern, where $X_2$ contributes only when added after $X_1$, and vice-versa. When two features are independent, no L-pattern is formed (Figure \ref{fig:Figure1}D). When two features are redundant, only the first added feature is capable of contributing substantially (Figure \ref{fig:Figure1}E). We  capture these relationships in the L-score metric, described below. 

\subsection*{L-Pattern Score: Mathematical Definition}

\[
\text{L\_score} = \frac{\text{skinny}_{\text{red}} \cdot \text{skinny}_{\text{blue}} \cdot (\text{horiz}_{\text{red}} - \text{horiz}_{\text{blue}})}{2}
\]

\textbf{Component Calculations:}
\begin{itemize}
    \item \textbf{Skinnyness} (for each point cloud):
    \[
    \text{skinny} = \frac{\lambda_1/\lambda_2}{1 + \lambda_1/\lambda_2}
    \]
    Where $\lambda_1$ and $\lambda_2$ are the primary and secondary PCA explained variance ratios. Values close to 1 indicate strong linear structure.

    \item \textbf{Horizontalness} (for each point cloud):
    \[
    \theta = (\arctan2(y,x)) \bmod 180 \\
    \quad \text{horiz} = \cos(2\theta)
    \]
    This maps angular orientation into a value between $-1$ (vertical) and $1$ (horizontal), with $0$ indicating diagonal structure.
\end{itemize}

\subsection*{Perfect L-Pattern}

In the ideal L-pattern, both point clouds are tightly aligned (high skinniness), and they are oriented along perpendicular axes, horizontal for one cloud and vertical for the other, for example:

\begin{align*}
\text{skinny}_{\text{red}}, 
\text{skinny}_{\text{blue}} &\approx 1 \\
\text{horiz}_{\text{red}} &\approx 1 \text{ (horizontal)} \\
\text{horiz}_{\text{blue}} &\approx -1 \text{ (vertical)}
\end{align*}

This corresponds to the geometric structure shown in Figure\ref{fig:Figure1}E: one feature $X_1$ accounts for most of the performance gain when added first (creating a horizontal red cloud), while the other (e.g., $X_2$) adds value only when added second (creating a vertical blue cloud). This asymmetry is captured in the sign and magnitude of the L-score:

\[
\text{L\_score} = \frac{1 \cdot 1 \cdot (-1 - 1)}{2} = -1 \quad )
\]

\[
\text{L\_score} = \frac{1 \cdot 1 \cdot (1 - (-1))}{2} = 1 \quad )
\]

The L-score is highest when both point clouds are elongated (skinny) and their orientations differ maximally. A positive score reflects redundancy, where either feature alone suffices Figure\ref{fig:Figure1}E. A negative score reflects synergy, where one feature is only useful in the presence of the other Figure\ref{fig:Figure1}C. A score near zero signifies independence Figure\ref{fig:Figure1}D

\subsection*{Implementation}

We implement the L-score methodology using two different strategies, depending on the desired balance between computational cost and completeness of feature interaction exploration.

\subsubsection*{Exhaustive Permutation-Based Approach}

In the exhaustive variant, we systematically explore all possible permutations of feature addition orders. For each permutation, we record the marginal reduction in mean squared error (MSE) achieved when a new feature is added into the model.

Specifically, for a given feature pair \((A, B)\), we track two types of events 
\begin{itemize}
    \item When \(A\) is added \emph{before} \(B\) in the sequence. In this case, the performance improvement from adding \(A\) is associated with the red cloud, since \(A\) is evaluated without prior influence from \(B\).
    \item When \(A\) is added \emph{after} \(B\). Here, the improvement is associated with the blue cloud, reflecting how \(A\)'s contribution is measured after \(B\) is already present in the model.
\end{itemize}

By exhaustively collecting these conditional performance reductions across all possible feature orderings, we construct a complete set of (x, y) points required to generate the L-pattern plots described earlier. This approach ensures that the clouds capture all interaction possibilities between features.

However, since the number of possible feature addition sequences grows factorially with the number of features (\(n!\) for \(n\) features), the exhaustive approach becomes computationally infeasible as \(n\) increases. Consequently, for larger feature sets, a more scalable, approximate method is required, motivating the development of a non-exhaustive (path-based) variant.

\subsubsection*{Non-Exhaustive (Path-Based) Approach}

In the path-based variant, we generate a controlled sample of random feature addition sequences rather than exhaustively exploring all permutations. For each trial, we generate a random permutation of all features. We then sequentially add each feature according to this random order, evaluating the model and recording the MSE reduction at each step. For each feature pair, we determine whether feature $A$ or $B$ was added first for each trial Figure\ref{fig:Figure1}B:

By conducting a sufficient number of random trials (typically scaling linearly with the feature count), we obtain a statistical approximation of the $(x, y)$ point clouds required for the L-pattern analysis. This sampling-based approach preserves the essential structure of feature interactions while dramatically reducing computational complexity from factorial ($n!$) to linear ($k \cdot n$, where $k$ is the number of random trials).

\newpage

\section{Synthetic Datasets and Validation Framework}

\subsection{Constructing Synergy-Based Datasets}

To evaluate the ability of our method to detect synergies between individual feature pairs, we construct synthetic datasets in which the target variable is a function of two interacting features. This design isolates pairwise relationships, enabling precise assessment of whether the L-score reflects synergy strength and direction. Additional features are included to serve as irrelevant distractors, allowing us to test the method's robustness in noisy, high-dimensional settings.

Each dataset consists of several input features, but only a specific pair, denoted $A$ and $B$, are involved in generating the output $Y$. The remaining features ($X_3$, $X_4$, etc.) are drawn from independent distributions and are unrelated to $Y$.

\subsubsection{Design of SynergyFunctions}

We define a variety of functional forms to represent different types of relationships between $A$ and $B$. These include:

\begin{itemize}
    \item \textbf{Multiplicative Synergy:} $Y = A \cdot B + \epsilon$\\
    A symmetric interaction where both features contribute equally and jointly.

    \item \textbf{Asymmetric Multiplicative Synergy:} $Y = A^3 \cdot B + \epsilon$\\
    An asymmetric formulation where one feature dominates due to a higher-order transformation.

    \item \textbf{Trigonometric Synergy:} $Y = \sin(A \cdot B) + \epsilon$\\
    A nonlinear relationship where interaction effects are preserved but harder to learn directly.
\end{itemize}

In all cases, $\epsilon$ is small Gaussian noise added to simulate natural variation or measurement error. The distractor features (e.g., $X_3$, $X_4$, $X_5$) are sampled independently from uniform or Gaussian distributions and do not contribute to the output, allowing us to test whether the L-score can ignore irrelevant information.

\subsection{Constructing Redundancy-Based Datasets}

To evaluate whether our method can detect redundancy, where multiple features encode overlapping information, we construct synthetic datasets in which several features are transformations of a shared underlying signal. This setup mimics real-world cases where features are correlated through latent structure or measurement artifacts, and the functional form of the redundancy may vary in strength and complexity.

We begin by generating a latent feature $A$ drawn from a continuous distribution, which serves as the sole driver of the output $Y$. Additional features are derived as transformations of $A$ and are therefore redundant with it to varying degrees. The target $Y$ is directly computed from $A$, meaning any predictive power in the other features arises purely from their shared origin.

We explore the following types of redundancy:

\begin{itemize}
    \item \textbf{Asymmetric Polynomial Redundancy:} $B = A^2 + \epsilon$\\
    A strong and smooth nonlinear relationship, fully preserving the sign and scale of the original signal.

    \item \textbf{Trigonometric Redundancy:} $C = \cos(\pi A) + \epsilon$\\
    A periodic transformation that introduces aliasing but retains a deterministic mapping to the latent variable.

    \item \textbf{Partial Redundancy (Lossy):} $D = |A| + \epsilon$\\
    A transformation that discards sign information, making the mapping non-invertible and only partially redundant.
\end{itemize}

 This design enables us to examine how the L-score responds to varying degrees of redundancy, including lossy transformations that obscure part of the underlying signal.

Code for producing all datasets and analysis is available at \url{https://github.com/A-Shamail/Ordering}

\section{Results}

\begin{figure}
    \centering
    \includegraphics[width=0.5\linewidth]{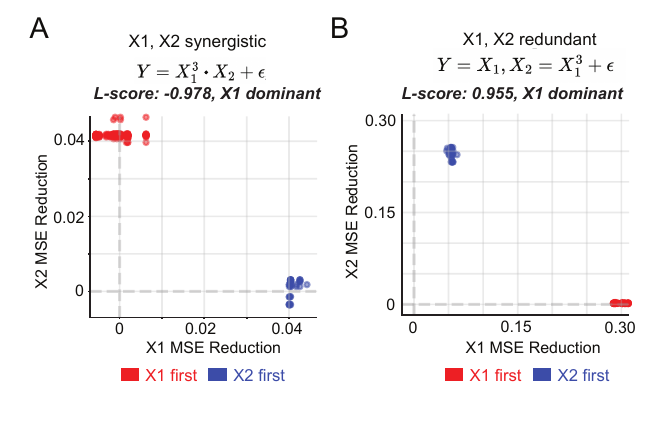}
    \caption{L-score scatter plot for the asymmetric synergy (A) and redundancy (B) datasets ($X_1$, $X_2$).  (A) For synergy The symmetric “L” structure indicates that neither feature alone is predictive, but the combination leads to substantial gain. (B) For redundancy, the first feature is most predictive, and the second adds little information. In both cases, the dominant feature has a higher average MSE reduction}
    \label{fig:Figure_synred}
\end{figure}

We demonstrate the L-score with two datasets 1) Asymmetric Polynomial Synergy, where $Y = X_1^3 \cdot X_2 + \epsilon$, and 2) Asymmetric Polynomial Redundancy, where $Y=X_1$ and $X_2 = X_1^3  + \epsilon$. In the first case, both $X_1$ and $X_2$ are essential to predict $Y$, though $X1$ is dominant to $X_2$ due to its higher order. In the second case, while $X_1$ and $X_2$ are both sufficient to predict $Y$, $X_1$ is the dominant predictor due to the noise component added to $X_2$. In both cases, we expect to find interaction between $X1$ and $X2$, and not between distractor features. 

For Asymmetric Polynomial Synergy (\ref{fig:Figure_synred}A), we see a strong negative L-score, characteristic of synergy. Skew between point clouds indicates feature dominance. The mean MSE values for $X_1$ are slightly higher than for $X_2$, showing that $X_1$ is the dominant feature in this pair. For Asymmetric Polynomial Redundancy (\ref{fig:Figure_synred}B), we see a strong positive L-score, characteristic of redundancy. In this case, the dominance of $X_1$ is much more pronounced, with an average MSE of $\sim$ 0.4, while $X_2$ has an average MSE of $\sim$ 0.35. 

\begin{figure}[h]
    \centering
    \includegraphics[width=1\linewidth]{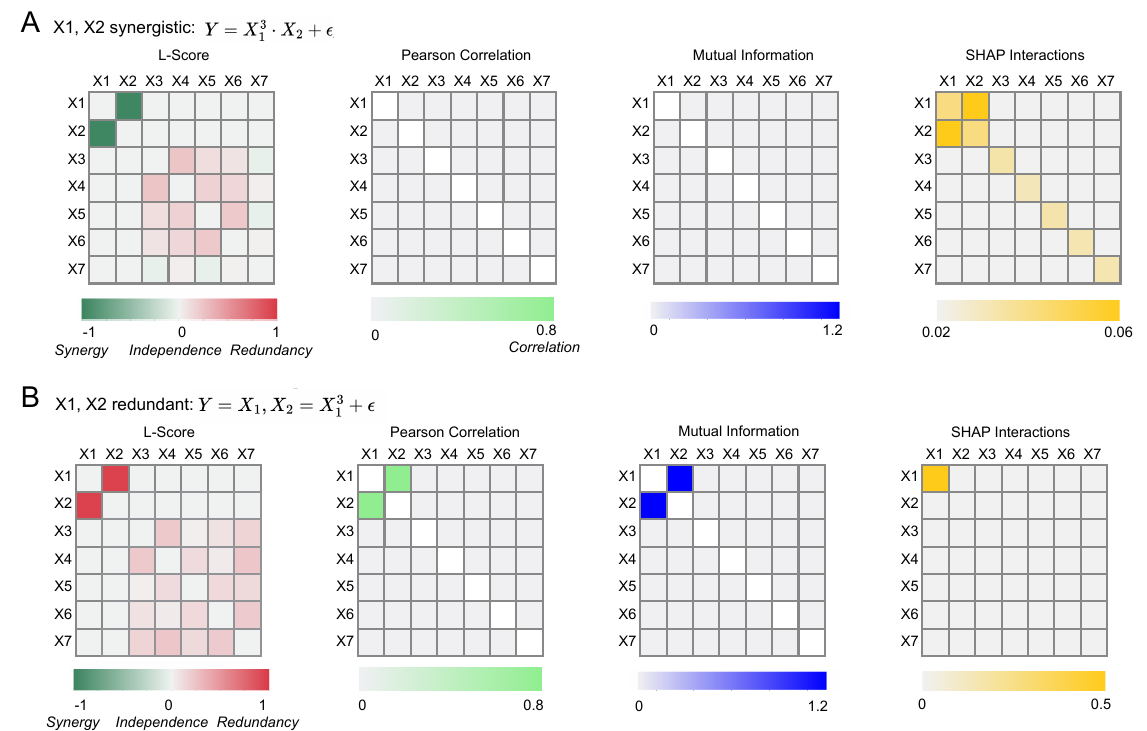}
    \caption{Comparison of correlation, mutual information, and SHAP values on the asymmetric synergy (A) and redundancy (B) datasets. (A) Only L-score and SHAP identify the synergistic interaction between $X_1$ and $X_2$. (B) Only L-score, Pearson Correlation, and Mutual Information identify the redundant relationship between $X_1$ and $X_2$}
    \label{fig:Figure_heatmaps}
\end{figure}

We next compare L-scores, Pearson Correlation, Mutual Information, and SHAP Interaction values for all pairs for features in these two datasets (\ref{fig:Figure_heatmaps}). When $X1$ and $X2$ interact synergistically, negative L-scores and positive SHAP Interaction values capture this relationship  (\ref{fig:Figure_heatmaps}A). As expected, Pearson Correlation and Mutual Information do not, as these features are created independently.  When $X_1$ and $X_2$ are redundant, positive L-scores, Pearson Correlation, and Mutual Information capture this relationship, while SHAP interactions do not (\ref{fig:Figure_heatmaps}B).

\subsection{Tri-Feature Dependency}

In tri-feature dependency, the target variable is a function of three features where no individual or pairwise combination is sufficient to predict the outcome. The dependency only becomes clear when all three are considered together. This reflects higher-order synergy between features.

\subsubsection*{Simple Example:}
\[
Y = X_1 \cdot X_2 \cdot X_3
\]

\begin{figure}[h]
    \centering
    \includegraphics[width=0.8\textwidth, height=15cm]{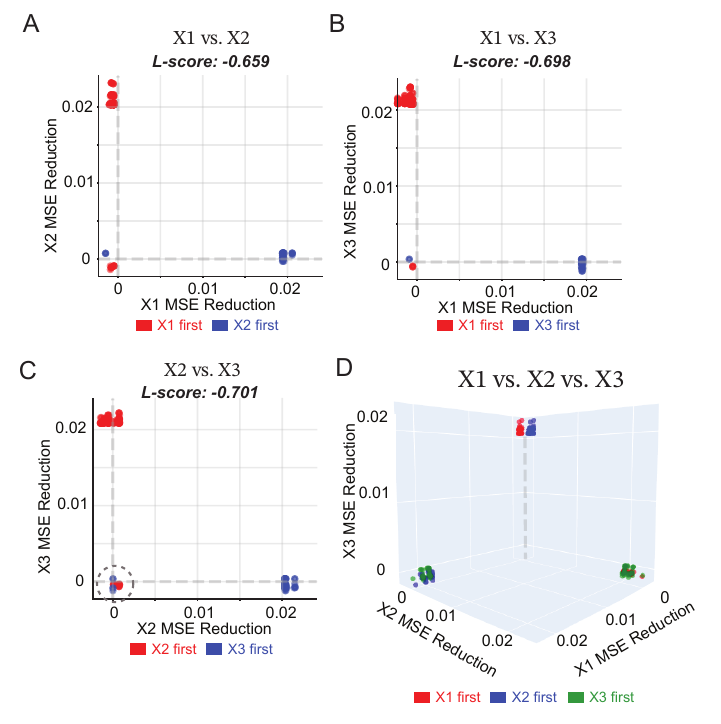}
    \caption{Tri-feature relationships, when $Y = X_1 \cdot X_2 \cdot X_3$ . In pairwise plots (A-C), points near the origin (circled in C) represent trials where the non-plotted feature was added last. All three features are plotted in D.}
    \label{fig:Figure3d}
\end{figure}

This multiplicative relationship ensures that none of the features alone, or in pairs, provide enough information to accurately predict $Y$. Only when all three features are jointly considered does the underlying structure reveal itself. This characteristic is clearly captured by our method (Figure~\ref{fig:Figure3d}). All feature pairs create strong L-score patterns of approximately $-0.7$.

The 3-dimensional plot ( Figure~\ref{fig:Figure3d} D reveals a compelling structure. In each case, the first and second features added contribute very little to model performance. The significant reduction in MSE only occurs when the third feature is introduced, regardless of the order in which the features are added. 

This consistent third-feature effect across permutations confirms that the predictive relationship among $X_1$, $X_2$, and $X_3$ is genuinely triadic, no individual or pair suffices. The information is only unlocked when the entire group is present. Importantly, pairwise L-scores capture this higher-order interaction triplet, without additional k-wise terms. 

\begin{table}[h]
\centering
\begin{tabular}{lcccc}
\toprule
\textbf{Relationship} & \textbf{L-Score}& \textbf{Mutual} & \multicolumn{2}{c}{\textbf{SHAP Analysis}} \\
\textbf{Type} & & \textbf{Information} & \textbf{Interaction} & \textbf{Individual Values} \\
\midrule
Symmetric Dependence & $\to -1$ & N/A & High & Low individual \\
(e.g., XY = Z) & (no skew) & (model-agnostic) & interaction & contributions \\
\addlinespace
Asymmetric Dependence & $\to -1$ & N/A & High & Dominant feature has \\
(e.g., X³Y = Z) & (skewed to X) & (model-agnostic) & interaction & higher contribution \\
\addlinespace
Independent & $\approx 0$ & Low & Low & Independent \\
(unique information) &  & & interaction & contributions \\
\addlinespace
Asymmetric Redundancy & $\to +1$ & High & Low & Dominant feature has \\
(e.g., X = Z, Y = |X|) & (skewed to X) & & interaction & higher contribution \\
\addlinespace
Symmetric Redundancy & $\to +1$ & High & Low & Similar \\
(e.g., X = 1/Y ) & (no skew) & & interaction & individual values \\
\bottomrule
\end{tabular}
\caption{Comparison of methods for feature relationship analysis. L-score values approach $-1$ for synergy and $+1$ for redundancy, with skew indicating asymmetry and the dominant feature. In $X^3Y = Z$, $X$ dominates due to its cubic effect; in $X = Z$, $Y = |X|$, $X$ dominates $Y$ by containing sign information lost in the absolute value. SHAP requires joint interpretation of interaction terms and individual values, while Mutual Information is model-agnostic and therefore symmetry-agnostic.}
\label{tab:method_comparison}
\end{table}

\section{Conclusion}

Understanding the nature of relationships between features, whether redundant, synergistic, or independent, is central to uncovering structure in data. In this work, we introduced a simple yet powerful idea based on random sequential feature addition, where model performance is tracked as features are incrementally revealed. From these empirical traces, we derive the \textbf{L-score}, a single interpretable metric that quantifies synergy, independence, and redundance between features on a unified $[-1,1]$ scale.

Through synthetic experiments, we demonstrated that the L-score successfully distinguishes symmetric redundancy from asymmetric dependency, identifies dominant contributors within related feature groups, and naturally extends to detect higher-order interactions. 

Table \ref{tab:method_comparison}.  highlights how the random-ordering framework provides a unified geometric view of feature relationships. Symmetric and asymmetric synergies both yield strongly negative L-scores, with skew revealing which feature contributes more strongly when interactions are asymmetric. Independent features cluster around an L-score of zero. Symmetric and asymmetric redundancy produce strongly positive L-scores, again with skew indicating dominance. Mutual Information aligns with this taxonomy by signaling shared information for redundant relationships, but cannot capture synergy or dominance. SHAP interaction values detect synergistic interactions and capture dominance through comparison of individual SHAP values, but do not detect redundancy. 

A distinctive advantage of the random-ordering approach is that pairwise L-scores naturally expose higher-order interaction structure without increasing combinatorial complexity. When three or more features jointly participate in synergistic or redundant relationships, the pairwise clouds (e.g., AB, AC, BC) exhibit coherent geometric patterns that reveal the underlying multi-element interaction.

Our method is simple to implement, compatible with arbitrary models, and scalable via randomized sampling. It provides not only a numeric signal of interaction strength and directionality but also geometric intuition through L-pattern plots.  This framework is extensible to any situation with discrete actions or features, and where error or progress can be tracked incrementally.  

In summary, random-ordering analysis and the L-score open a path toward detecting non-obvious interaction structure in complex systems. Beyond standard machine-learning settings, this framework could detect subtle biological interactions, expose functional relationships in symbolic regression tasks, and guide the discovery of informative and non-redundant feature sets for ensemble methods. By providing a unified, interpretable, and model-agnostic measure of synergy and redundancy, the method offers a foundation for deeper understanding of how components work together in high-dimensional domains.



\newpage
\bibliographystyle{plain}
\bibliography{references}

@article{friedman2008predictive,
  title={Predictive learning via rule ensembles},
  author={Friedman, Jerome H and Popescu, Bogdan E},
  journal={The Annals of Applied Statistics},
  volume={2},
  number={3},
  pages={916--954},
  year={2008},
  publisher={Institute of Mathematical Statistics},
  doi={10.1214/07-AOAS148}
}

@misc{apley2019visualizingeffectspredictorvariables,
      title={Visualizing the Effects of Predictor Variables in Black Box Supervised Learning Models}, 
      author={Daniel W. Apley and Jingyu Zhu},
      year={2019},
      eprint={1612.08468},
      archivePrefix={arXiv},
      primaryClass={stat.ME},
      url={https://arxiv.org/abs/1612.08468}, 
}

@misc{owen2013higherordersobolindices,
      title={Higher order Sobol' indices}, 
      author={Art Owen and Josef Dick and Su Chen},
      year={2013},
      eprint={1306.4068},
      archivePrefix={arXiv},
      primaryClass={math.NA},
      url={https://arxiv.org/abs/1306.4068}, 
}

@misc{lundberg2017unifiedapproachinterpretingmodel,
      title={A Unified Approach to Interpreting Model Predictions}, 
      author={Scott Lundberg and Su-In Lee},
      year={2017},
      eprint={1705.07874},
      archivePrefix={arXiv},
      primaryClass={cs.AI},
      url={https://arxiv.org/abs/1705.07874}, 
}

@misc{dhamdhere2020shapleytaylorinteractionindex,
      title={The Shapley Taylor Interaction Index}, 
      author={Kedar Dhamdhere and Ashish Agarwal and Mukund Sundararajan},
      year={2020},
      eprint={1902.05622},
      archivePrefix={arXiv},
      primaryClass={cs.GT},
      url={https://arxiv.org/abs/1902.05622}, 
}

@misc{tsai2023faithshapfaithfulshapleyinteraction,
      title={Faith-Shap: The Faithful Shapley Interaction Index}, 
      author={Che-Ping Tsai and Chih-Kuan Yeh and Pradeep Ravikumar},
      year={2023},
      eprint={2203.00870},
      archivePrefix={arXiv},
      primaryClass={cs.LG},
      url={https://arxiv.org/abs/2203.00870}, 
}







\end{document}